\documentclass[11pt]{article}
\usepackage{paclic33}
\usepackage{times}
\usepackage{latexsym}
\usepackage{amsmath}
\usepackage{multirow}
\usepackage{url}

\usepackage{kotex}
\usepackage{graphicx}
\usepackage[normalem]{ulem}
\useunder{\uline}{\ul}{}

\usepackage{floatpag}

\title{Investigating an Effective Character-level Embedding \\in Korean Sentence Classification}

\author{Won Ik Cho, Seok Min Kim, and Nam Soo Kim\\
	Human Interface Laboratory\\
	Department of Electrical and Computer Engineering and INMC\\
	Seoul National University\\ 1 Gwanak-ro, Gwanak-gu, Seoul, Korea, 08826\\
	{\tt \{wicho,smkim\}@hi.snu.ac.kr, nkim@snu.ac.kr}\\
}

\date{}

\begin{document}
\maketitle
\begin{abstract}
  Different from the writing systems of many Romance and Germanic languages, some languages or language families show complex conjunct forms in character composition. For such cases where the conjuncts consist of the components representing consonant(s) and vowel, various character encoding schemes can be adopted beyond merely making up a  one-hot vector. However, there has been little work done on intra-language comparison regarding performances using each representation. In this study, utilizing the Korean language which is character-rich and agglutinative, we investigate an encoding scheme that is the most effective among \textit{Jamo}\footnote{Letters of Korean alphabet \textit{Hangul}.}-level one-hot, character-level one-hot, character-level dense, and character-level multi-hot. Classification performance with each scheme is evaluated on two corpora: one on binary sentiment analysis of movie reviews, and the other on multi-class identification of intention types. The result displays that the character-level features show higher performance in general, although the \textit{Jamo}-level features may show compatibility with the attention-based models if guaranteed adequate parameter set size. 
\end{abstract}

\section{Introduction}

Ever since an early approach exploiting the character features for the neural network-based natural language processing (NLP) ~\cite{zhang2015character},  character-level embedding\footnote{Throughout this paper, the terms \textit{embedding} and \textit{encoding} are parallelly used depending on the context.} has been widely used for many tasks such as machine translation ~\cite{ling2015character}, noisy document representation ~\cite{dhingra2016tweet2vec}, language correction ~\cite{xie2016neural}, and word segmentation ~\cite{cho2018real}. However, little consideration was done for intra-language performance comparison regarding variant representation types. Unlike English, a Germanic language written with an alphabet comprising 26 characters, many languages used in East Asia are written with scripts whose characters can be further decomposed into sub-parts representing individual consonants or vowels. This conveys that (sub-)character-level representation for such languages has the potential to be managed with more than just a simple one-hot encoding.

In this paper, a comparative experiment is conducted on Korean, a representative language with a featural writing system ~\cite{daniels1996world}. To be specific, the Korean alphabet \textit{Hangul} consists of the letters \textit{Jamo} denoting consonants and vowels. The letters comprise a morpho-syllabic block that refers to \textit{character}, which is resultingly equivalent to the phonetic unit \textit{syllable} in terms of Korean morpho-phonology. The conjunct form of a character is \{Syllable: CV(C)\}; this notation implies that there should be at least one consonant (namely \textit{cho-seng}, the first sound) and one vowel (namely \textit{cwung-seng}, the second sound) in a character. An additional consonant (namely \textit{cong-seng}, the third sound) is auxiliary.  However, in decomposition of the characters, three slots are fully held to guarantee a space for each component; an empty cell comes in place of the third entry if there is no auxiliary consonant. The number of possible sub-characters, or (composite) consonants/vowels, that can come for each slot is 19, 21, and 27. For instance, in a syllable `간 (\textit{kan})', the three clock-wisely arranged characters ㄱ, ㅏ, and ㄴ, which sound \textit{kiyek} (stands for \textit{k}; among 19 candidates), \textit{ah} (stands for \textit{a}; among 21 candidates), and \textit{niun} (stands for \textit{n}; among 27 candidates), refers to the \textit{first}, the \textit{second} and the \textit{third} sound respectively.

To compare five different \textit{Jamo}/character-level embedding methodologies that are possible in Korean, we first review the related studies and the previous approaches. Then, two datasets are introduced, namely binary sentiment classification and multi-class intention identification, to investigate the performance of each representation under recurrent neural network (RNN)-based analysis. After searching for an effective encoding scheme, we demonstrate how the result can be adopted in combating other tasks and discuss if a similar approach can be applied to the languages with the complex writing system.

\section{Related Work}

Inter-language comparison with word and character embedding was thoroughly investigated in ~\newcite{zhang2017encoding}, for Chinese, Japanese, Korean, and English. The paper investigates the languages via representations including \textit{character, byte, romanized character, morpheme}, and \textit{romanized morpheme}. The observation of tendency for Korean suggests that adopting the raw characters outperforms utilizing the romanized character-level features, and moreover both the performance are far beyond the morpheme-level features. However, to be specific on the experiment, decomposition of the morpho-syllabic blocks was not conducted, and the experiment did not make use of the dense embedding methodologies which can project the distributive semantics onto the representation. We concluded that more attention is to be paid to different character embedding methodologies of Korean. Here, to reduce ambiguity, we denote a morpho-syllabic block which consists of consonant(s) and a vowel by \textit{character}, and the individual components by \textit{Jamo}. 
A \textit{Jamo} sequence is spread in the order of the \textit{first} to the \textit{third} sound if a \textit{character} is decomposed.

There has been little study done on an effective text encoding scheme for Korean, a language that has distinguished character structure which can be decomposed into sub-characters. 
A comparative study on the hierarchical constituents of Korean morpho-syntax was first organized in ~\newcite{lee2016comparison}, in the way of comparing the performance of \textit{Jamo, character, morpheme}, and \textit{eojeol (word)}-level embeddings for the task of text reconstruction. In the quantitative analysis using edit distance and accuracy, the \textit{Jamo}-level feature showed a slightly better result than the character-level one. The (sub-)character-level representations presented the outcome far better than the morpheme or \textit{eojeol}-level cases, as in the classification task of ~\newcite{zhang2017encoding}. The results show the task-independent competitiveness of the character-level features.

In a more comprehensive viewpoint, ~\newcite{stratos2017sub} showed that \textit{Jamo}-level features combined with word and character-level ones display better performance with the parsing task.
With more elaborate character processing, especially involving \textit{Jamo}s, ~\newcite{shin2017korean} and ~\newcite{cho2018character} made progress recently in the classification tasks.
~\newcite{song2018sequence} aggregated the sparse features into multi-hot representation successfully, enhancing the output within the task of error correction. In a slightly different manner, ~\newcite{cho2018real} applied dense vectors for the representation of the characters, obtained by skip-gram ~\cite{mikolov2013distributed}, improving the naturalness of word segmentation for noisy Korean text. To figure out the tendency, we implement the aforementioned \textit{Jamo}/character-level features and discuss the result concerning classification tasks. The details on each approach are to be described in the following section.

\begin{table*}[]
	\centering
	\resizebox{0.9\textwidth}{!}{%
		\begin{tabular}{|c|c|c|c|c|}
			\hline
			\textbf{} & \textbf{Representation} & \textbf{Property} & \textbf{Dimension} & \textbf{Feature type} \\ \hline
			\textbf{(i) \textit{Shin2017}} & ㄱ $\cdots$ ㅎ / ㅏ $\cdots$ ㅢ / ㄱ $\cdots$ ㅄ & \textit{Jamo}-level & 67 & one-hot \\ \hline
			\textbf{(ii) \textit{Cho2018c}} & (i) + ㄱ $\cdots$ ㅏ $\cdots$ ㅄ & \textit{Jamo}-level & 118 & one-hot \\ \hline
			\textbf{(iii) \textit{Cho2018a}-Sparse} & $\cdots$ 간 $\cdots$ 밤 $\cdots$ 핫 $\cdots$ & character-level & 2,534 & one-hot \\ \hline
			\textbf{(iv) \textit{Cho2018a}-Dense} & $\cdots$ 간 $\cdots$ 밤 $\cdots$ 핫 $\cdots$ & character-level & 100 & dense \\ \hline
			\textbf{(v) \textit{Song2018}} & $\cdots$ 간 $\cdots$ 밤 $\cdots$ 핫 $\cdots$ + $\alpha$ & character-level & 67 & multi-hot \\ \hline
		\end{tabular}%
	}
	\caption{A description on the \textit{Jamo}/character-level features (i-v).}
	\label{my-label}
\end{table*}

\section{Experiment}

In this section, we demonstrate the featurization of five (sub-)character embedding methodologies, namely (i) \textbf{\textit{Jamo}} ~\cite{shin2017korean,stratos2017sub} (ii) \textbf{modified \textit{Jamo}} ~\cite{cho2018character}, (iii) \textbf{sparse \textit{character} vectors}, (iv) \textbf{dense \textit{character} vectors} ~\cite{cho2018real} trained based on fastText ~\cite{bojanowski2016enriching}, and (v) \textbf{multi-hot \textit{character} vectors} ~\cite{song2018sequence}. We featurize only  \textit{Jamo}/character and no other symbols such as numbers and special letters is taken into account.

For (i), we used one-hot vectors of dimension 67 (= 19 + 21 + 27), which is smaller in width than the ones suggested in ~\newcite{shin2017korean} and ~\newcite{stratos2017sub}, due to the omission of special symbols. Similarly, for (ii), 118-dim one-hot vectors are constructed. The different point of (ii) regarding (i) is that it considers the cases that \textit{Jamo} is used in the form of single (or composite) consonant or vowel, as frequently observed in the social media text. The cases make up an additional array of dimension 51. 

For (iii) and (iv), we adopted a vector set that is distributed publicly in ~\newcite{cho2018real}, reported to be extracted from a drama script corpus of size 2M. 
Constructing the vectors of (iii) is intuitive; for $N$ characters in the corpus, a $N$-dimensional one-hot vector is assigned for each. Case of (iv) can be considered awkward in the sense of using characters as a meaningful token, but we concluded that the Korean characters can be handled as a word piece\footnote{The word piece models were not investigated in this study since here we concentrate on the (sub-)character-level embeddings.} ~\cite{sennrich2015neural} or subword n-gram ~\cite{bojanowski2016enriching} concerning the nature of their composition. All the characters are reported to be treated as a separate token (subword) in the training phase that uses skip-gram ~\cite{mikolov2013distributed}. 
Although the number of possible character combinations in Korean is precisely 11,172 (= 19 * 21 * 28), the number of ones that are used in real-life reaches about 2,500 ~\cite{kwon1995contextual}. Since the description says that the corpus is removed with punctuation and consists of around 2,500 Korean syllables, we exploited the dictionary of 100-dim fastText-embedded vectors which is provided in the paper, and extracted the list of the characters to construct a one-hot vector dictionary\footnote{Two types of embeddings were omitted, namely the \textit{Jamo}-based fastText and the 11,172-dim one-hot vectors; the former was considered inefficient since there are only 118 symbols at most and the latter was assumed to require a huge computation.}. 

(v) is a hybrid of \textit{Jamo} and character-level features;  three vectors indicating the first to the third sound of a character, namely the ones with dimension 19, 21, and 27 each, are concatenated into a single multi-hot vector. This preserves the conciseness of the \textit{Jamo}-level one-hot encodings and also maintains the characteristics of conjunct forms.
In summary, (i) utilizes 67-dim one-hot vectors, (ii) 118-dim one-hot vectors, (iii) 2,534-dim one-hot vectors, (iv) 100-dim dense vectors, and (v) 67-dim multi-hot vectors (Table 1).

\subsection{Task description}

For evaluation, we employed two classification tasks that can be conducted with the character-level embeddings. Due to a shortage of reliable open source data for Korean, we selected the datasets that show a clear description of the annotation. One, a sentiment analysis corpus, is expected to display how well each character-level encoding scheme conveys the information regarding lexical semantics. The other, an intention analysis corpus, is expected to show how comprehensively each character-level encoding scheme deals with the syntax-semantic task that concerns sentence form and content. The details on each corpus are stated below.

\subsubsection{Naver sentiment movie corpus}

The corpus NSMC\footnote{https://github.com/e9t/nsmc} is a widely used benchmark for evaluation of Korean language models. The annotation follows ~\newcite{maas2011learning} and incorporates 150K:50K documents for the train and test set each. The authors assign a positive label for the reviews with a score $>$ 8 and negative for the ones with a score $<$ 5 (in 10-scale), adopting a binary labeling system. To prevent confusion that comes from gray-zone data, neutral reviews were removed. The positive and negative instances are equally distributed in both train and test set. 

\subsubsection{Intonation-aided intention identification for Korean}

The corpus 3i4K\footnote{https://github.com/warnikchow/3i4k} ~\cite{cho2018speech} is a recently distributed open-source data for multi-class intention identification. The labels, in total seven, include \textit{fragment} and five clear-cut cases (\textit{statement, question, command, rhetorical question (RQ), rhetorical command (RC)}). The remaining class is for the intonation-dependent utterances whose intention mainly depends on the prosody assigned to underspecified sentence enders, considering head-finality of the Korean language. Since the labels are elaborately defined and the corpus is largely hand-labeled (or hand-generated), the corpus size is relatively small (total 61K) and some classes possess a tiny volume (e.g., about 1.7K for RQs and 1.1K for RCs).  However, such challenging factors of the dataset can show the aspects of the evaluation that can be overlooked. The train-test ratio is 9:1.

\subsection{Feature engineering}

In the first task, to manage with the document size, the length of \textit{Jamo} or character sequence was fixed to the maximum of (i-ii) 420 and (iii-v) 140\footnote{The data description says the maximum volume of the input characters is 140.}. Similarly, in the second task, (i-ii) 240 and (iii-v) 80\footnote{The number of the utterances with the length longer than 80 were under 40 ($<0.07\%$).}. The length regarding (i-ii) being three times as long as that of (iii-v) comes from the spreadings of the sub-characters for each character.

For both tasks, the document was numericalized in the way that the tokens are placed on the right end of the feature, to preserve \textit{Jamo}s or characters which may incorporate syntactically influential components of the phrases in a head-final language. For example, in a sentence ``배고파 (pay-ko-pha, \textit{I'm hungry})'', a vector sequence is arranged in the form of [0 0 $\cdots$ 0 0 $v1$ $v2$ $v3$], where $v1$, $v2$, and $v3$ each denotes the vector embeddings of the characters \textit{pay}, \textit{ko}, and \textit{pha}. Here, \textit{pha} encompasses the head of the phrase with the highest hierarchy in the sentence, which assigns the sentence a speech act of \textit{statement}. The spaces between \textit{eojeol}s were represented as zero vector(s)\footnote{\textit{Eojeol} denotes the unit of spacing in the written Korean.}. 

To look into the content of the corpora, the first dataset (NSMC) contains many incomplete characters such as solely used sub-characters (e.g., ㅋㅋ, ㅠㅠ) and non-Korean symbols (e.g., Chinese characters, special symbols, punctuations). The former ones were treated as characters, whereas the latter ones were ignored in all features. Although (i, iii, iv) do not represent the symbols regarding the former as non-zero vector while (ii, v) do so, we concluded that this does not threaten the fairness of the evaluation, since a wider range of representation is own advantage of each feature. The second dataset (3i4K) contains only the full characters. Thus no disturbance or biasedness was induced in the featurization.


\begin{table*}[]
	\centering
	\resizebox{0.8\textwidth}{!}{%
		\begin{tabular}{|c|c|c|c|c|}
			\hline
			\multirow{2}{*}{\textbf{\begin{tabular}[c]{@{}c@{}}Accuracy\\ (F1-score)\end{tabular}}} & \multicolumn{2}{c|}{\textbf{NSMC}} & \multicolumn{2}{c|}{\textbf{3i4K}} \\ \cline{2-5} 
			& \textit{BiLSTM} & \textit{BiLSTM-SA} & \textit{BiLSTM} & \textit{BiLSTM-SA} \\ \hline
			\textbf{(i) \textit{Shin2017}} & 0.8203 & 0.8316 & 0.8694 (0.7443) & \textbf{0.8769} (0.7692) \\ \hline
			\textbf{(ii) \textit{Cho2018c}} & 0.7895 & 0.7973 & 0.8688 (0.7488) & 0.8728 (0.7727) \\ \hline
			\textbf{(iii) \textit{Cho2018a}-Sparse} & \textbf{0.8271} & \textbf{0.8321} & 0.8694 {\ul (0.7763)} & 0.8722 (0.7741) \\ \hline
			\textbf{(iv) \textit{Cho2018a}-Dense} & \textbf{0.8312} & \textbf{0.8382} & \textbf{0.8799} {\ul (0.7887)} & \textbf{0.8844} {\ul (0.7963)} \\ \hline
			\textbf{(v) \textit{Song2018}} & \textbf{0.8271} & 0.8314 & \textbf{0.8696} (0.7713) & 0.8761 {\ul (0.7828)} \\ \hline
		\end{tabular}%
	}
	\caption{Performance comparison. Only the accuracy is provided for NSMC since the labels are equally distributed. Two best models regarding accuracy (and F1-score for 3i4K) are bold (and underlined) for both tasks, with each architecture (BiLSTM and BiLSTM-SA).}
	\label{my-label}
\end{table*}

\subsection{Implementation}

The implementation was done with Hangul Toolkit\footnote{https://github.com/bluedisk/hangul-toolkit}, fastText\footnote{https://pypi.org/project/fasttext/}, and Keras \cite{chollet2015keras}, which were used for character decomposition, dense vector embedding and RNN-based training, respectively. For RNN models, bidirectional long short-term memory (BiLSTM) ~\cite{schuster1997bidirectional} and self-attentive sentence embedding (BiLSTM-SA) ~\cite{lin2017structured} were applied. 

In vanilla BiLSTM, an autoregressive system that is representatively utilized for time-series analysis, a fully connected layer (FC) is connected to the last hidden layer of BiLSTM, finally inferring the output with a softmax activation.
In BiLSTM with a self-attentive embedding, the context vector whose length equals to that of the hidden layers of the BiLSTM, is jointly trained along with the network so that it can provide the weight assigned to each hidden layer. The weight is obtained by making up an attention vector via a column-wise multiplication of the context vector and the hidden layers. The model specification is provided as supplementary material.

\subsection{Result}

For both tasks, we split the train set into 9:1 to have a separate validation set. As a result, we achieved 135K instances for the training of NSMC (15K for the validation) and 50K for the training of 3i4K (5K for the validation).

\subsubsection{Performance}

The evaluation phase displays the pros and cons of the conventional methodologies (Table 2). In both tasks, (iv) showed significant performance. It is assumed that the result comes from the distinguished property of (iv); it does not break up the syllabic blocks and at the same time provides the distributional semantics to the models, in the way of employing skip-gram ~\cite{mikolov2013distributed}. (v) also performs in a similar way, by using a multi-hot encoding that assigns own role to each vector representation, displaying a compatible performance using BiLSTM in both tasks. 

(iii) preserves the blocks as well, but one-hot encoding hardly gives any information on each character. It is assumable that such representation can be powerful for the dataset with a rich and balanced resource, as in NSMC, but is weak if the class volume is imbalanced, which led to an insignificant result for 3i4K. Although some compatible performance was achieved with BiLSTM, the models regarding (iii) reached saturation fast and displayed overfitting afterward, while the models with the other features showed a gradual increase in accuracy. The reason for fast saturation seems to be the limited flexibility coming from the vast parameter set size. 

The unexpected point is that the models utilizing additional letters (ii) showed significant performance degeneration in NSMC task, where the solely used sub-characters (as ㅋㅋ implying joy or ㅠㅠ implying sadness) were expected to be aggregated into the featurization and yield a positive outcome. In the pilot research executed without validation set (that the model performing best with the test set was searched greedily), a comparable result as in (i) was shown. Thus, the reason for the degeneration seems to be the limitation of using a validation set, where the cutback in the training resource is inevitable\footnote{It is highly recommended to use the cross-validation if one wants to boost the performance.}. Also, some solely used sub-characters might have caused the disorder in the inference of the sentiment, since not all the users employ the sentiment-related sub-characters in the same way. Supporting this observation, feature (ii) shows much less difference with (i) in 3i4K, where only the full characters are adopted.

\subsubsection{Using self-attentive embedding}

The advantage of using self-attentive embedding was the most emphasized in \textit{Jamo}-level feature (i) for both tasks, and the least in (iii) (Table 2). 
We assume that relatively more significant improvement using (i) originates in the decomposability of the blocks. If a sequence of the blocks is decomposed into the sequence of sub-characters, the morphemes can be highlighted to provide more syntactically/semantically meaningful information to the system, especially the ones that could not have been revealed in the block-preserving environment (iii-v). For example, a Korean word `이상한 (\textit{i-sang-han}, strange)' can be split into `이상하 (\textit{i-sang-ha}, the root of the word)' and `-ㄴ (\textit{-n}, a particle that makes the root an adjective)', making up the circumstances in which the presence and role of the morphemes is pointed out. 
This property is also reflected in the case of using the feature (ii), although the absolute score is not notable.

\subsubsection{Decomposability vs. Local semantics}

The point described above is the disadvantage of character-level features (iii-v) in the sense that in such ones, characters cannot be decomposed, even for the sparse multi-hot encoding. The higher performance of (iv-v) compared to the \textit{Jamo}-level features, which is currently displayed, can hence be explained as a result of preserving the cluster of letters. If the computation resource is sufficient so that exploiting deeper networks (e.g., Transformer ~\cite{vaswani2017attention} or BERT ~\cite{devlin2018bert}) is available, we infer that (i-ii) may also show compatible or better performance, since the modern self-attention-based mechanisms utilize the positional encodings to grasp the relation between the tokens, advanced from the location-based models we adopted. Nevertheless, it is still quite impressive that (iv) scores the highest even though the utilized dictionary does not incorporate all the available character combinations. It is suspected to be where the distributive semantics on the word pieces are engaged in. 

\begin{table}[]
	\centering
	\resizebox{\columnwidth}{!}{%
		\begin{tabular}{|c|c|c|c|c|}
			\hline
			\multirow{2}{*}{\textbf{\begin{tabular}[c]{@{}c@{}}Trainable \\ param.s \&\\ Training time\end{tabular}}} & \multicolumn{2}{c|}{\textbf{\textit{BiLSTM}}} & \multicolumn{2}{c|}{\textbf{\textit{BiLSTM-SA}}} \\ \cline{2-5} 
			& Param.s & \begin{tabular}[c]{@{}c@{}}Time /\\ epoch\end{tabular} & Param.s & \begin{tabular}[c]{@{}c@{}}Time /\\ epoch\end{tabular} \\ \hline
			\textbf{(i) \textit{Shin2017}} & 34,178 & 13.5m & 297,846 & 18m \\ \hline
			\textbf{(ii) \textit{Cho2018c}} & 47,234 & 16m & 310,902 & 20.5m \\ \hline
			\textbf{(iii) \textit{Cho2018a}-Sparse} & 665,730 & 33m & 772,318 & 38.5m \\ \hline
			\textbf{(iv) \textit{Cho2018a}-Dense} & 42,626 & 6.5m & 149,214 & 6m \\ \hline
			\textbf{(v) \textit{Song2018}} & 34,178 & 6m & 140,766 & 6m \\ \hline
		\end{tabular}%
	}
	\caption{Computation burden for NSMC models.}
	\label{my-label}
\end{table}

\subsubsection{Computation efficiency}

In this study, we investigate only on the classification tasks. Notwithstanding they take a short amount of time for training and inference, the measurement on parameter volume and complexity is meaningful (Table 3). It is observed that (v) yields a compatible or better performance with respect to the other schemes, accompanied by less burden of computation. Besides, we argue that the multi-hot encoding (v) has a significant advantage over the rest in terms of multiple usages; it possesses both conciseness of the sub-character (\textit{Jamo})-level features and local semantics (although not distributional) of the character-level features. Due to these reasons, the derived models are fast in training and also have potential to be effectively used in sentence reconstruction or generation, as shown in ~\newcite{song2018sequence}, where applying large-dimensional one-hot encoding has been considered challenging.

\section{Discussion}

The primary goal of this paper is to search for a \textit{Jamo}/character-level encoding scheme that best resolves the given task in Korean NLP. Empirically, we found out that the fastText-embedded vectors outperform the other features if provided with the same environment (model architecture). It is highly probable that the distributive semantics plays a significant role in the NLP tasks concerning syntax-semantics, at least in the feature-based approaches ~\cite{mikolov2013distributed,pennington2014glove}.
However, we experimentally verified that even with traditional feature-based systems, the sparse encoding schemes also perform adequately with the dense one, especially displaying computation efficiency in the multi-hot case. 

At this moment, we want to emphasize that the utility of the comparison result is not only restricted to Korean, in that the introduced character encoding schemes are also available in other languages. Although the Korean writing system is unique, the Japanese language incorporates several morae (e.g., small \textit{tsu}) that approximately correspond to the third sound (\textit{cong-seng}) of the Korean characters, which may let the Japanese characters be encoded in a similar manner with the cases of Korean. Also, each character of the Chinese language (and \textit{kanji} in Japanese) can be further decomposed into sub-characters (\textit{bushu} in Japanese) that have meanings as well, as suggested in ~\newcite{nguyen2017sub} (e.g., 鯨 ``whale" to 魚 ``fish" and 京 ``capital city").

Besides, many other languages that are used in South Asia (India), such as Telugu, Devanagari, Tamil, and Kannada, have writing system type of Abugida\footnote{https://en.wikipedia.org/wiki/Writing\_system} ~\cite{daniels1996world}, the composition of consonant and vowel. The cases are not the same as Korean in view of a writing system since featural decomposition of the Abugida characters is not represented in the way of segmentation of a glyph. However, for example, instead of listing all the CV combinations, one can simplify the representation by segmenting the property of the character into consonant and vowel and making up a two-hot encoded vector. The similar kind of character embedding can be applied to many native Philippine languages such as Ilocano. Moreover, we believe that the argued type of featurization is robust in combating the noisy user-generated texts.


\section{Conclusion}

In this study, we have reviewed the five different types of (sub-)character-level embedding for a character-rich language. It is remarkable that the dense and multi-hot representation perform best given the classification tasks, and specifically, the latter one has the potential to be utilized beyond the given tasks due to its conciseness and computation efficiency. The utility of the sub-character-level features is also noteworthy in the syntax-semantic tasks that require morphological decomposition. It is expected that the overall performance tendency may provide a useful reference for the text processing of other character-rich languages with conjunct forms in the writing system, including Japanese, Chinese, and the languages of various South and Southeast Asian regions. 
A brief tutorial on both datasets using embedding methodologies presented in this paper is available online\footnote{https://github.com/warnikchow/kcharemb}.

\section*{Acknowledgement}

This research was supported by Projects for Research and Development of Police science and Technology under Center for Research and Development of Police science and Technology and Korean National Police Agency funded by the Ministry of Science, ICT and Future Planning (PA-J000001-2017-101). Also, this work was supported by the Technology Innovation Program (10076583, Development of free-running speech recognition technologies for embedded robot system) funded by the Ministry of Trade, Industry \& Energy (MOTIE, Korea). The authors appreciate Yong Gyu Park for giving helpful opinions in performing validation and evaluation. After all, the authors want to send great thanks to the three anonymous reviewers for the insightful comments.

%

%
%
%
%
%
%

\bibliography{my_bib_190624}
\bibliographystyle{acl}

\newpage

\onecolumn
\section*{\centering Supplementary Material}
\subsection*{\centering \textit{BiLSTM} and \textit{BiLSTM-SA} model specification} 

\subsection*{Variables}
\begin{itemize}
	\item Sequence length (L) and the number of output classes (N) depend on the task. For NSMC, L = 420 for feature (i-ii) and 140 for (iii-v). For 3i4K, L=240 for feature (i-ii) and 80 for (iii-v). N equals 2 and 7 for NSMC and 3i4K, respectively.
	\item Character vector dimension (D) depends on the feature. For features (i-v), D equals 67, 118, 2534, 100, and 67, respectively.
\end{itemize}

\subsection*{BiLSTM}
\begin{itemize}
	\item Input dimension: (L, D)
	\item RNN Hidden layer width: 64 (=32$\times$2)
	\item The width of FCN connected to the last hidden layer: 128 (Activation: \textit{ReLU})
	\item Output layer width: N (Activation: \textit{softmax})
\end{itemize}

\subsection*{BiLSTM-SA}
\begin{itemize}
	\item Input dimension: (L, D)
	\item The dimension of RNN hidden layer sequence output: (64 (= 32$\times$2), L) \\$>>$ each layer connected to FCN of width: 64 (Activation: \textit{tanh}; equals to $d_a$ in ~\newcite{lin2017structured}) [a]
	\item Auxiliary zero vector size: 64 \\$>>$ connected to FCN of width 64 (Activation: \textit{ReLU}, Dropout ~\cite{srivastava2014dropout}: 0.3) \\$>>$ connected to FCN of width 64 (Activation: \textit{ReLU}) [b]
	\item Vector sequence [a] is column-wisely dot-multiplied with [b] to yield the layer of length L
	\\$>>$ connected to an attention vector of size L (Activation: \textit{softmax}) \\$>>$ column-wisely multiplied to the hidden layer sequence to yield a weighted sum [c] of width 64 
	\\$>>$ [c] is connected to an FCN of width: 256 (Activation: \textit{ReLU}, Dropout: 0.3) $\times$ 2 (multi-layer)
	\item Output layer width: N (Activation: \textit{softmax})
\end{itemize}

\subsection*{Settings}
\begin{itemize}
	\item Optimizer: Adam ~\cite{kingma2014adam} (Learning rate: 0.0005)
	\item Loss function: Categorical cross-entropy
	\item Batch size: 64 for NSMC, 16 for 3i4K (due to the difference in the corpus volume)
	\item For 3i4K, class weights were taken into account to compensate the volume imbalance.
	\item Device: Nvidia Tesla M40 24GB
\end{itemize}

\end{document}